%% file: main.tex
\documentclass[letterpaper, 10 pt, conference]{ieeeconf}  % Comment this line out
\IEEEoverridecommandlockouts                              % This command is only
\usepackage{booktabs}
\usepackage{url}
\usepackage[usenames,dvipsnames,svgnames,table]{xcolor}
\usepackage{amsmath} % assumes amsmath package installed
\usepackage{amssymb}  % assumes amsmath package installed
\usepackage{pifont}

\usepackage{subcaption}
\usepackage{multirow}
\usepackage{multicol}

\newcommand{\captionstyle}{\sf \footnotesize }

\usepackage{listings,xcolor}
\usepackage{todonotes}
\usepackage{xspace}

\title{\LARGE \bf Balancing Shared Autonomy with Human-Robot Communication}

\author{Rosario Scalise, Yonatan Bisk, Maxwell Forbes, Daqing Yi, Yejin Choi, and Siddhartha Srinivasa\\
Paul G. Allen School of Computer Science and Engineering, University of Washington\\
\small \texttt{\{rosario, ybisk, mbforbes, dqyi, yejin, siddh\}@ cs.washington.edu}}% <-this % stops a space
\begin{document}
% https://tex.stackexchange.com/questions/336919/simple-coloring-of-json-attributes
\lstset{
    string=[s]{``}{''},
    stringstyle=\color{blue},
    comment=[l]{:},
    commentstyle=\color{black},
}

\maketitle
\thispagestyle{empty}
\pagestyle{empty}

%%%%%%%%%%%%%%%%%%%%%%%%%%%%%%%%%%%%%%%%%%%%%%%%%%%%%%%%%%%%%%%%%%%%%%%%%%%%%%%%
\begin{abstract}
Robotic agents that share autonomy with a human should leverage human domain knowledge and account for their preferences when completing a task. This extra knowledge can dramatically improve plan efficiency and user-satisfaction, but these gains are lost if  communicating with a robot is taxing and unnatural.  In this paper, we show how
viewing human-robot language through the lens of shared autonomy explains
the efficiency versus cognitive load trade-offs humans make when deciding how cooperative and explicit to make their instructions. 
\end{abstract}

%%%%%%%%%%%%%%%%%%%%%%%%%%%%%%%%%%%%%%%%%%%%%%%%%%%%%%%%%%%%%%%%%%%%%%%%%%%%%%%%
\section{Introduction} 
Human-Robot Interaction research often focuses on constructing fully-autonomous systems that work with humans or language-based interfaces that take explicit commands as a form of linguistic teleoperation.  Since both the robot and the human have reasoning abilities and unique capabilities, we propose that language can form the basis of a system with shared autonomy wherein both agents help each other accomplish a task by leveraging their individual strengths.  In this paper, we investigate how constraints from language balance the cognitive loads both the human and robot need to handle to complete the task, and the language's robustness when scaling the problem size.

Our work views incorporating language with robots through the lens of shared autonomy~\cite{javdani2015shared}. 
Autonomy falls on a spectrum ranging from fully autonomous systems (e.g., manufacturing robots) that require no input from humans, to simple devices (e.g., telescopic cranes) that rely entirely on the user to specify the next action. All scenarios require both reasoning and action. Often we assume both tasks should fall to the robot, but shared autonomy aims to share the burden across all participants.  This also allows the user to bias actions of their robotic assistant.  This work presents a series of user studies on how humans naturally communicate about a simple table-clearing task to their robot butler, and how knowledge of robotic manipulation capabilities, linguistic comprehension, and task complexity affect a user's language.

\begin{figure}[t!]
    \centering
    \includegraphics[width=0.95\linewidth]{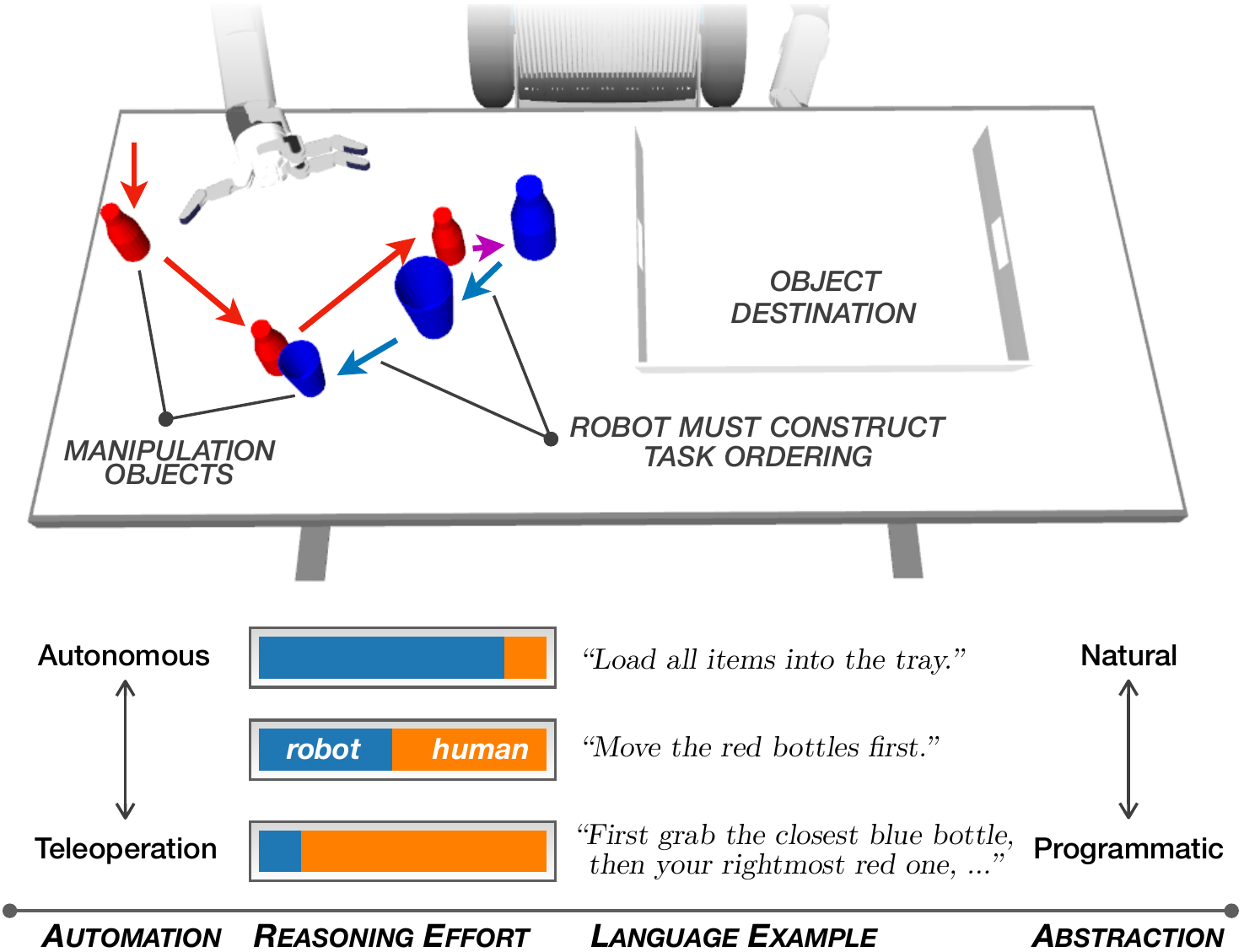}
    \caption{\captionstyle Shared autonomy is a spectrum along which efforts of both the human and robot trade off in service of a task. We show that a human's expectations of the robot's agency manifest in their utterances.}
    \label{fig:spectrum}
\end{figure}

Complex environments make planning difficult and automatic plans do not account for user preferences.
For example, when packing a suitcase, one might want their toiletries on top for easy access, and they will need to communicate this constraint.  When asked for input on plan creation, we find that humans are fast and effective planners, but only once they understand the robot's capabilities.  In linguistics, the nature of language a person produces is believed to be guided by Grice's Maxims \cite{frederking1996grice}.  We paraphrase two of them here:  

\vspace{5pt}
\begin{itemize}
\item[]\hspace{-25pt} \textbf{Quantity:} A speaker tries to be as informative as possible while giving only as much information as is necessary.
\item[]\hspace{-25pt} \textbf{Manner:} A speaker tries to be brief and to avoid ambiguity.
\end{itemize}
\vspace{5pt}

What is most important to understand about these principles is that they are fundamentally social.  To know what or how much to say requires modeling the listener.  In the case of robotics, most users have either a weak or incorrect model of the robot's abilities and so, without guidance, they make the wrong assumptions about what is necessary and what is useful to communicate.  We explore this later in the paper.

We hypothesize that language varies from natural to programmatic (Fig. \ref{fig:spectrum}) depending on a user's expectations of the robots capabilities.  Our second hypothesis is that if asked to, humans are effective and efficient planners whose insights, preferences and task intuitions can be effectively harnessed to improve task performance.  Our experiments use the task and motion planning library, MAGI~\cite{srinivasa2016system}, to translate the high-level table-clearing task into an ordering of table-top objects and the corresponding motion plans needed to execute moving each object.\footnote{The planner performs a depth-first search over a permutation graph of manipulable objects until it finds a geometrically feasible solution which addresses all objects.} We setup our experiments within this planning framework to allow users to specify high-level goals or orders of objects to manipulate.  We avoid bias that might be introduced by working with members of the university community by running studies on Amazon's Mechanical Turk (AMT) with participants who had no previous knowledge of robotics.  We constrain the experiments to a simple table clearing task \cite{srinivasa2016system} but feel these results generalize to other domains and robotic platforms.  

We first describe a preliminary study (\S \ref{sec:prelim}) which guides the creation of our AMT study on comparing plan efficiency and user effort (\S \ref{sec:exp1}).  Next, we investigate a new type of language to the literature that occurs in complex domains (\S \ref{sec:exp2}) and analyze the language strategies users produced  (\S \ref{sec:lang_strategy}).

\section{Related Work}
Our work is motivated by the rich breadth of research on using natural language to communicate with robots both within the robotics and natural language communities.
We are interested in how language interaction shifts the cognitive load in shared autonomy.

Previous work has grounded natural language navigational commands to executable representations. 
Graphical model-based approaches using syntactic parses have been applied to controlling robotic forklift actions \cite{tellex2011understanding} and mobile navigation in novel environments \cite{hemachandra2015learning,yi2016expressing}. 
Others have utilized CCG semantic parsing of robotic commands in synthetic environments \cite{matuszek2013learning}, and with weak supervision \cite{artzi2013weakly}. 
Howard (2014) \cite{howard2014natural} parsed natural language to constraints in trajectory space in order to reduce the search space of their graphical model that grounds language to instructions. Park (2017) \cite{park2017generating} used a similar approach, grounding instructional language to a probabilistic graphical model, though they address the manipulation domain and learn soft cost functions to avoid dynamically-specified regions (e.g., ``don't put it there''). 

Language interaction provides useful information in solving many of the challenges in shared autonomy, which include how to correctly and accurately identify a human's intent through interaction ~\cite{javdani2015shared} and observation~\cite{dragan2013legibility}.  
Low-level control strategies from humans have also been successfully integrated and applied to search terrain~\cite{okada2011shared}. Recent research aggressively incorporates high-level human information in shared autonomy~\cite{ahmed2013bayesian,yi2017incorporating}. Language, which is a direct and natural way for a human to share information in collaboration, has been widely researched within the ``supervisor" paradigm \cite{kollar2013toward,matuszek2012joint,howard2014natural} where only a goal is provided or where the human acts as a ``programmer" \cite{sato1987language,lauria2002mobile,forbes2015robot} that instructs the agent in tedious detail.  Balancing these roles and investigating user preferences is still an open challenge for deploying communication in shared autonomy.

Finally, when referencing ambiguous objects, people use visual attributes and referring expressions to indicate which object should move next.  Like our work, approaches to referring expressions largely focus on the tabletop domain \cite{matuszek2012joint}, and offer systems that are interactive \cite{kollar2013toward}, collaborative \cite{fang2015embodied}, or incorporate manipulation actions \cite{forbes2015robot}. 
Referring expressions can also be used as a means of effectively asking for help \cite{tellex2014asking}. 
Finally, the Natural Language Processing (NLP) community has recently also introduced corpora for this task with substantially more complex linguistic constructions \cite{bisk2016natural}.  Given such varied language between papers, our work aims to better understand what leads a user to choose a given communicative strategy.

\section{Preliminary Study}
\label{sec:prelim}
Our preliminary study investigates whether humans are good planners and how their language changes when they are taught about a robot's capabilities and planner.  The training we provide partway through the study is a proxy for the online learning they would receive when interacting with a robot, but allows us to collect multiple data points for an individual user. This provides us a clear before-and-after comparison to investigate the effect of training. 

We explained the goal task (table-clearing) to the user and that our robot, HERB, will only be using his right arm to manipulate one object at a time (partially replicated below). We then provided ten object configurations images like those shown in the top box of Figure \ref{fig:configs} (the design rational of which is explained in greater detail in section \ref{sec:studydesign}) to nine participants and requested instructions one-by-one:
\begin{small}
\begin{quote}
    \textit{You are overseeing a robot (HERB)’s role as a butler, ensuring he completes his tasks correctly and efficiently. Occasionally, he needs a little guidance.}\\
    ...\\
    \textit{For each scenario below, write an instruction for HERB to follow.  Remember, all cups must end up in the tray.}
\end{quote}
\end{small}

\begin{figure}[t!]
    \centering
    \begin{subfigure}[t]{0.48\linewidth}
    \includegraphics[width=\linewidth]{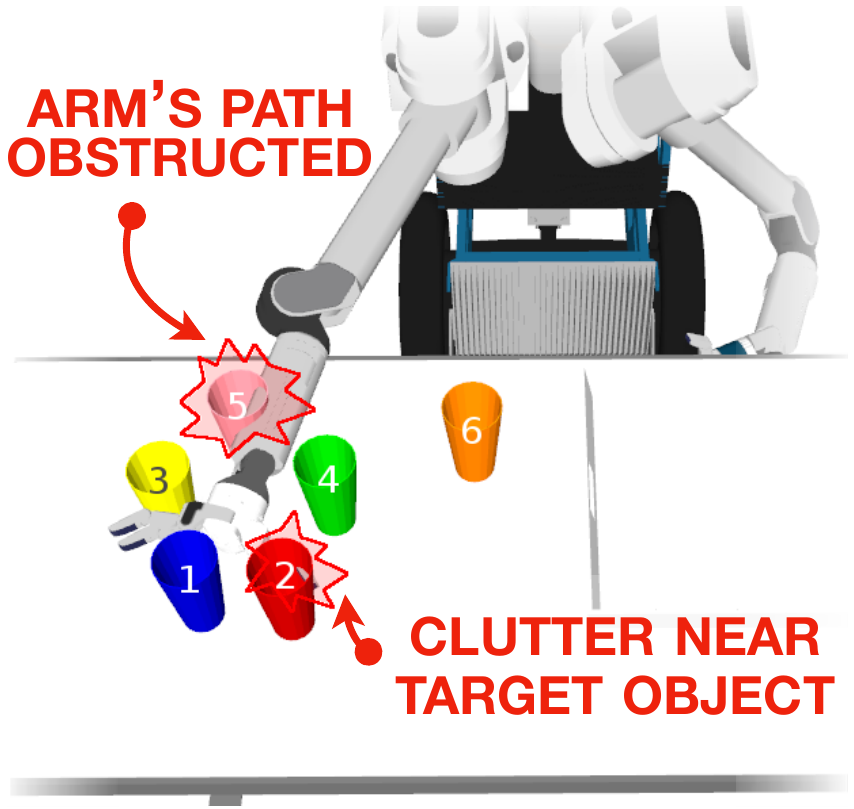}
    \caption{\captionstyle Low-level motion planning might not find a collision-free solution to a multi-step manipulation task if forced to follow a bad action order.}
    \label{fig:training}
    \end{subfigure}
    ~
    \begin{subfigure}[t]{0.48\linewidth}
    \includegraphics[width=\linewidth]{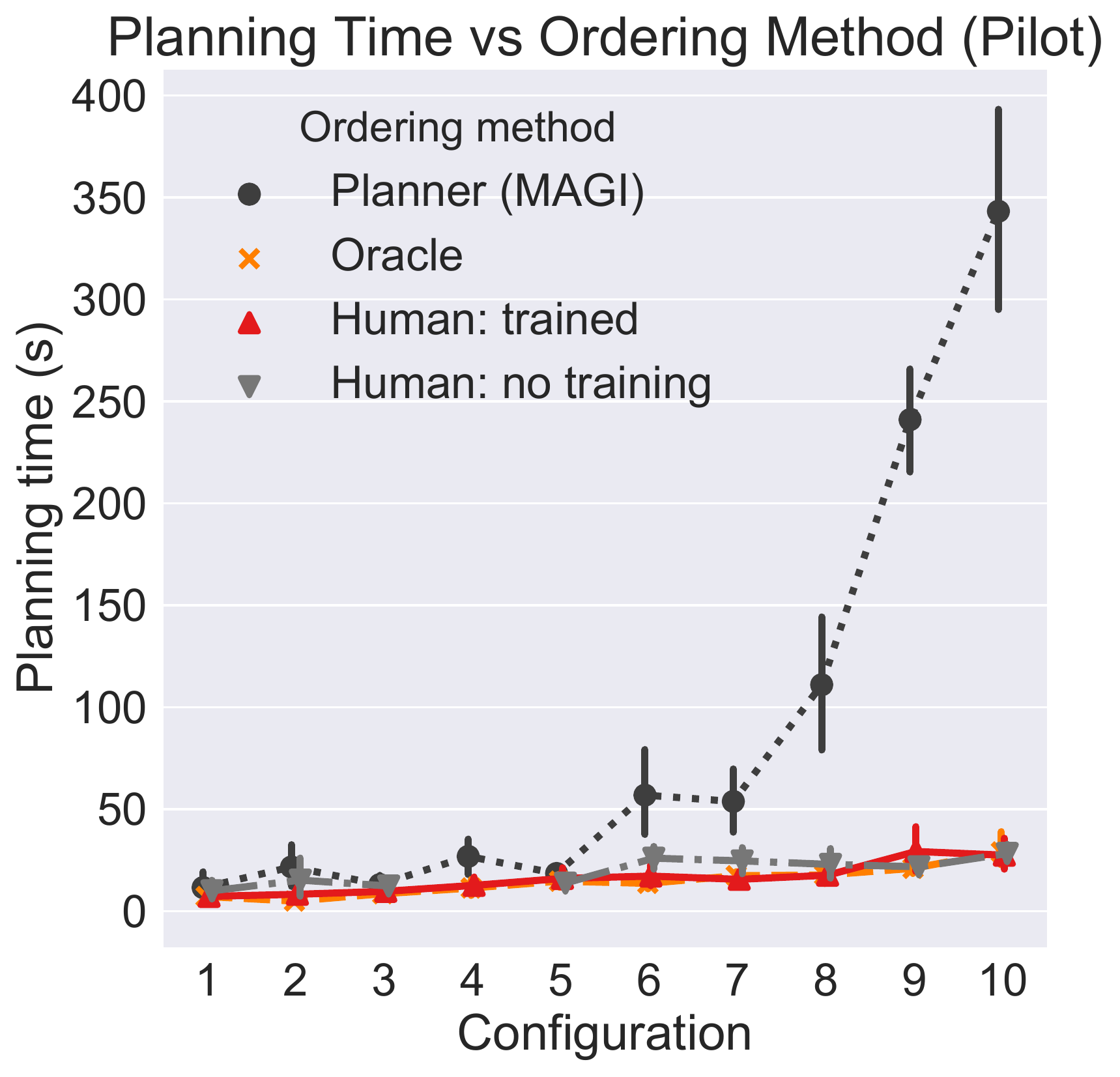}
    \caption{\captionstyle Planning time increases significantly when the task is complex,
    but incorporating human insight can offer dramatic performance gains.}
    \label{fig:pilot}
    \end{subfigure}
\end{figure}

\subsection{Untrained}  For the initial experiment no additional information was provided.  Unsurprisingly, most participants provided vague goal-oriented language:
\begin{enumerate}
    \item \textit{Place all cups in the tray} \hfill \begin{footnotesize}\texttt{[goals]}\end{footnotesize}
    \item \textit{Pick up the closest cup and move it onto the tray.  Repeat until there are no more cups.} \hfill \begin{footnotesize}\texttt{[algorithmic]}\end{footnotesize}
\end{enumerate}

Even a user who anticipated the fine-grained planning needs of the robot and initially provided verbose step-by-step instructions like: \textit{``Pick up the red cup and put it in the tray, then pick up the blue cup and put it in the tray, then...''} eventually switched to vague phrasings after growing bored with the task. 

As it is clear that these ``plans" contain no immediately useful information for the planner, once all scenarios were completed we asked participants to ``translate" their own instructions into an explicit ordering over subtasks corresponding to each object on the table.  Here, they are making concrete their own assumptions about how to complete the task.  We pass these orderings to the planner and compare its planning times to those produced by vanilla planner runs (meaning it has no priors on ordering over subtasks for objects on the table).  Plan performances are shown in Figure \ref{fig:pilot}. The ten instructions cover tables with three through seven uniquely colored cups for both densely and loosely cluttered scenes.  We chose these to compare the breadth of difficulty for the planner as compared to the human participants.

\subsection{Training}
Next, we train users on basic details of planning algorithms in colloquial terms and with a technical definition:
\begin{small}
\begin{quote}
    \textit{HERB must sample and evaluate trajectories to assess if he can reach the cup and manipulate it.  The fewer viable paths, the more samples are rejected and the harder it is to plan successfully.}
\end{quote}
\end{small}

Alongside this explanation we provide the user a basic demonstration of what to avoid when giving an ordering (Figure \ref{fig:training}). 
We do not instruct users in how to communicate.
After training, we repeat our data collection with ten new prompts.  This brief training leads participants to converge on the use of long verbose plans with explicit orderings.  

\subsection{Preliminary Experimental Results}
This preliminary study shows that a small amount of training dramatically affects robot-oriented language in a way that's easier for the robot but less natural and more cumbersome for the person (quantitative metrics and analysis of our large-scale studies are presented in section \ref{sec:exp1}).  Next we evaluate the plans users produced to see how well human intuitions map to good motion plans. 

Figure \ref{fig:pilot} suggests two important findings.  First, humans intuit good orderings well (outperforming our baseline in planning time), and second, even untrained users perform well.  This might indicate that their initial high-level instructions assumed a very capable planner, and when training introduced doubts they made their assumptions explicit.

\section{Study design}
\label{sec:studydesign}

\subsection{Stimulus Design}
We programatically generated stimulus images which consisted of colored cups (and/or) bottles in varying configurations on a table top. Each of the images was rendered from a 45 degree inclination angle and had the robot, HERB, present sitting in a pre-task configuration behind the table from the viewers perspective. Each of the configurations could be varied by the number of objects in the scene, whether the objects were sparsely placed or packed tightly, and by the object attributes available (object type, color, and size). When scaling the number of objects in a configuration, we chose to incrementally add an additional object to an existing set to create the new configuration. Doing this allowed us to be confident that there were no additional factors that could affect the changing complexity between two configurations.
\\

\begin{figure}[t!]
    \centering
    \includegraphics[width=\linewidth]{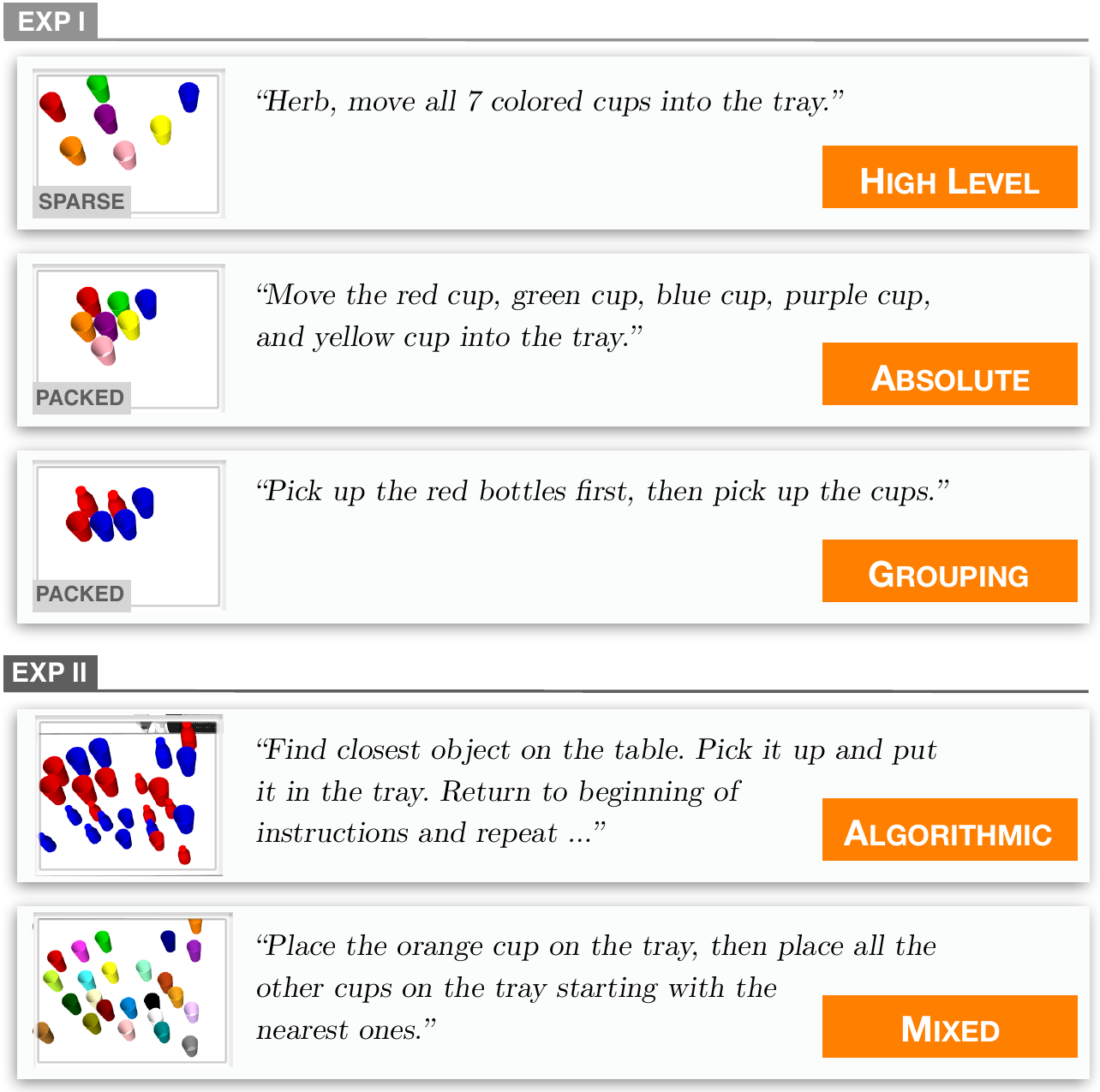}
    \caption{\captionstyle Example configurations and language from both experiments. The orange boxes denote the type of language used in the shown example.}
    \label{fig:configs}
\end{figure}

\noindent \textbf{Experiment I Stimuli}
Our first experiment consists of two types of stimuli (Figure \ref{fig:configs}).
The first set of stimulus images was composed of 5 `sparse' configurations and 5 `packed' configurations (each of which contained between 3 and 7 uniquely colored large-sized cups) for a total of 10 images.

The second set of images was composed of 6 `sparse' configurations and 6 `packed' configurations where the number of objects for all configurations was fixed at 6, colors were restricted to only red and blue (no longer unique), and objects' attributes were systematically varied. We generated a `sparse' and a `packed' configuration for each possible combination of available discriminating object attributes. For example, in the first two configurations, color is the only available attribute while in the last two configurations, color, type, and size are all varied among the objects. Our rational for picking 6 objects and not more was that this resulted in sufficient complexity without creating longer run-times for our planner. In total, the second set contained 12 images.
\\

\noindent \textbf{Experiment II Stimuli}
Our second experiment investigates the effect of scaling and complexity.  Here 
the set of stimulus images was composed of 6 configurations with 24 uniquely colored cups and 6 configurations of 24 objects in which they were randomly assigned attributes (a color from ${red,blue}$, a size from ${small,large}$, and a type from ${cup,bottle}$). 
In total, the third set contained 12 images (examples in Fig.~\ref{fig:configs}).

\subsection{Subject Allocation}

Both Experiments I and II were deployed via AMT. We recruited a total of n=50 participants for each, ensuring that participants who had seen Experiment I were not eligible to do Experiment II. We required that each participant was a native English speaker and was not color blind. We also surveyed participants on their past experience with robots at the end of each study.

\section{Experiment I}
\label{sec:exp1}

The results of the preliminary study corroborate our intuition that researching human robot collaboration in decision making falls within the framework and goals of shared autonomy. Specifically, we noted that humans are good at high-level reasoning and can specify an efficient ordering over subtasks in a high-level planning task. However, this shifts the cognitive workload to the human which they are reluctant to accommodate.  In contrast, robots are unlikely to have context specific heuristics about the environment and therefore have more difficulty finding a good orderings over subtasks, but they are very good at low-level path-planning for generating motion trajectories for individual subtasks and are highly capable assistants. This asymmetry between humans and robots can be generalized to a wide range of shared autonomy systems. Both agents will always have different capabilities or specialties, and shifting the cognitive load required by the human biases the system in one direction on the spectrum of full-autonomy to full teleoperation. 

Understanding the breadth of language people tend to use in these interactions helps to
\begin{enumerate}
    \item design language understanding algorithms and language communication schemes to support human-robot communication;
    \item enable a robot to interpret the current intent of a human in sharing the workload; and 
    \item allow a robot to actively generate language in order to shift its contribution on the autonomy spectrum to optimize the team performance.
\end{enumerate}

Our first experiment is designed to test two hypotheses:

\noindent \textbf{H1} \emph{Humans tend to provide ``natural" expressions when they trust in the robot's capabilities (they assume high robot autonomy before any information is given), which requires less work (mental + temporal demands).}

\noindent \textbf{H2} \emph{Humans tend to provide ``programmatic" expressions when they are aware of a robot's limitations, despite it increasing their workload.}  

\subsection{Study Design}
\label{sec:exp1_design}
We use the stimuli discussed in \S \ref{sec:studydesign} in our experiment on AMT.  Training is split into two phases: 1. \textbf{R}: Robot capabilities and 2. \textbf{R+C}: Information about language phrasings that the robot can understand (constraints and orderings).  This two phase approach ensures that our ``translation" step from the preliminary study cannot bias participants towards listing objects. 

\subsection{Evaluation}
To test our hypotheses we perform three evaluations based on plan efficiency, participant self-assessment, and linguistic analysis.\\
\\
\noindent \textbf{Plan Efficiency}
We annotate all plans provided by users into orderings (or partial-orderings) over the manipulable objects in the scene. With these orderings, we run the motion planner and assess each plan's efficiency by averaging over 10 plan times. When users provide high-level instructions (which contain no ordering information), we use the planner's de facto performance on the task. When users provide instructions which map to partial-orderings, we average the planning time of 3 samples from the set of feasible orderings which satisfy the given partial ordering.  \\
\\
\noindent \textbf{Participant Self-Assessment}
We ask our participants two categories of question: 1. We ask about the effort involved in the task and 2. We ask if they prefer scenes with simple referents or sets of objects.
User effort is collected with questions based on NASA TLX~\cite{hart2006nasa}. For each instruction we ask:
\begin{enumerate}
    \item \textit{How mentally demanding was it to come up with this?}
    \item \textit{How much time did it take you to think of this?}
    \item \textit{How satisfied are you with your performance on this?}\\
\end{enumerate}

\noindent \textbf{Linguistic Analysis}
We introduce two metrics for how the language of our participants vary through several conditions. First, we analyze type-token ratios.  Natural language is highly varied and rarely formulaic.  This manifests in a high type-token ratios (number of unique word types vs number of total words used).   In contrast, when language becomes repetitive and programmatic the ratio will fall.  We therefore present type-token ratios for all conditions as a proxy for how programmatic and ``unnatural" language has become.

Second, we manually annotated all of the responses as falling into one of three categories: \textit{high-level}, \textit{partial ordering}, and \textit{absolute ordering}.  We therefore report the number of instructions in each condition which belong to these classes.

\subsection{Results}
Our primary interest is in understanding the trade-off between the mental energy exerted by the participant versus their informativeness to the planner.

%
% Language figures: for both Experiment 1 and 2
%
\begin{figure}
    \centering

    \begin{subfigure}[t]{0.48\linewidth}
    \includegraphics[width=\linewidth]{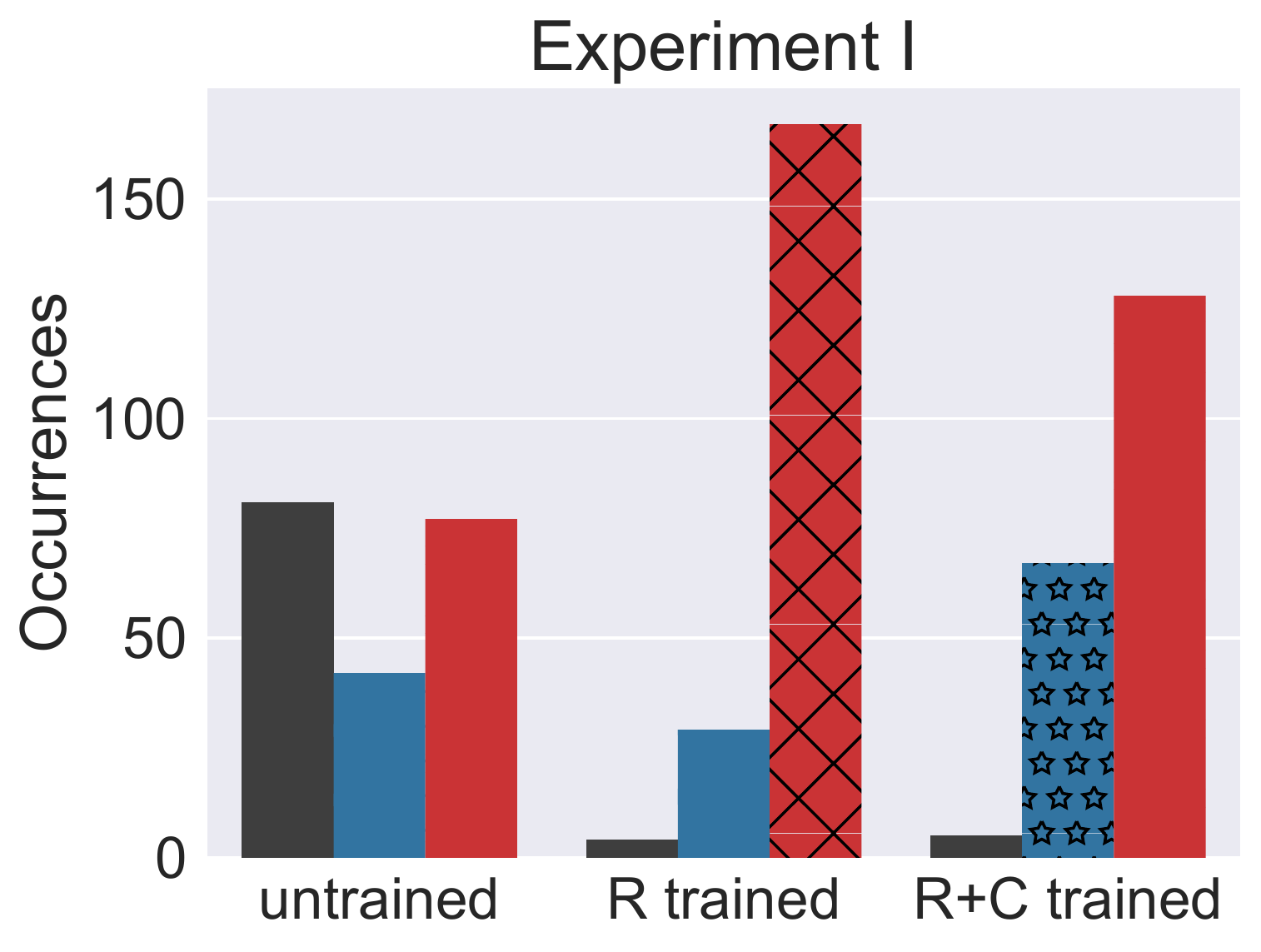}
    \label{fig:exp1_ordering}
    \end{subfigure}
    ~
    \centering
    \begin{subfigure}[t]{0.48\linewidth}
    \includegraphics[width=\linewidth]{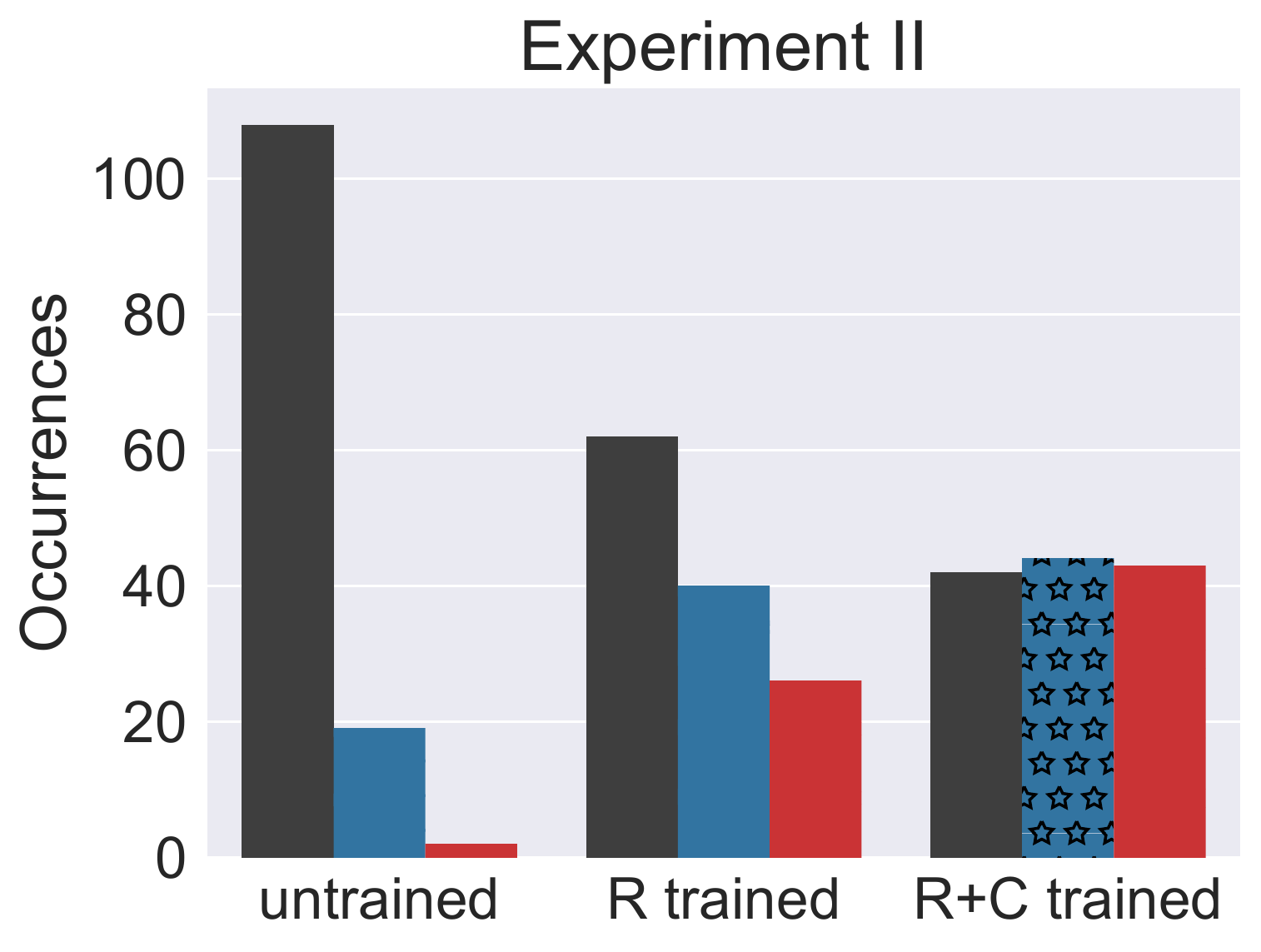}
    \label{fig:exp2_ordering}
    \end{subfigure}

    \vspace{-.3cm}

    \includegraphics[width=0.7\linewidth]{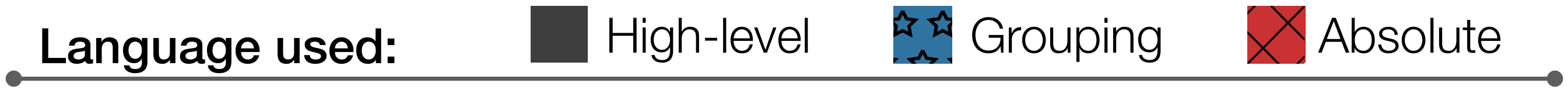}
    
    \vspace{.5cm}

    \begin{subfigure}[t]{0.48\linewidth}
    \includegraphics[width=\linewidth]{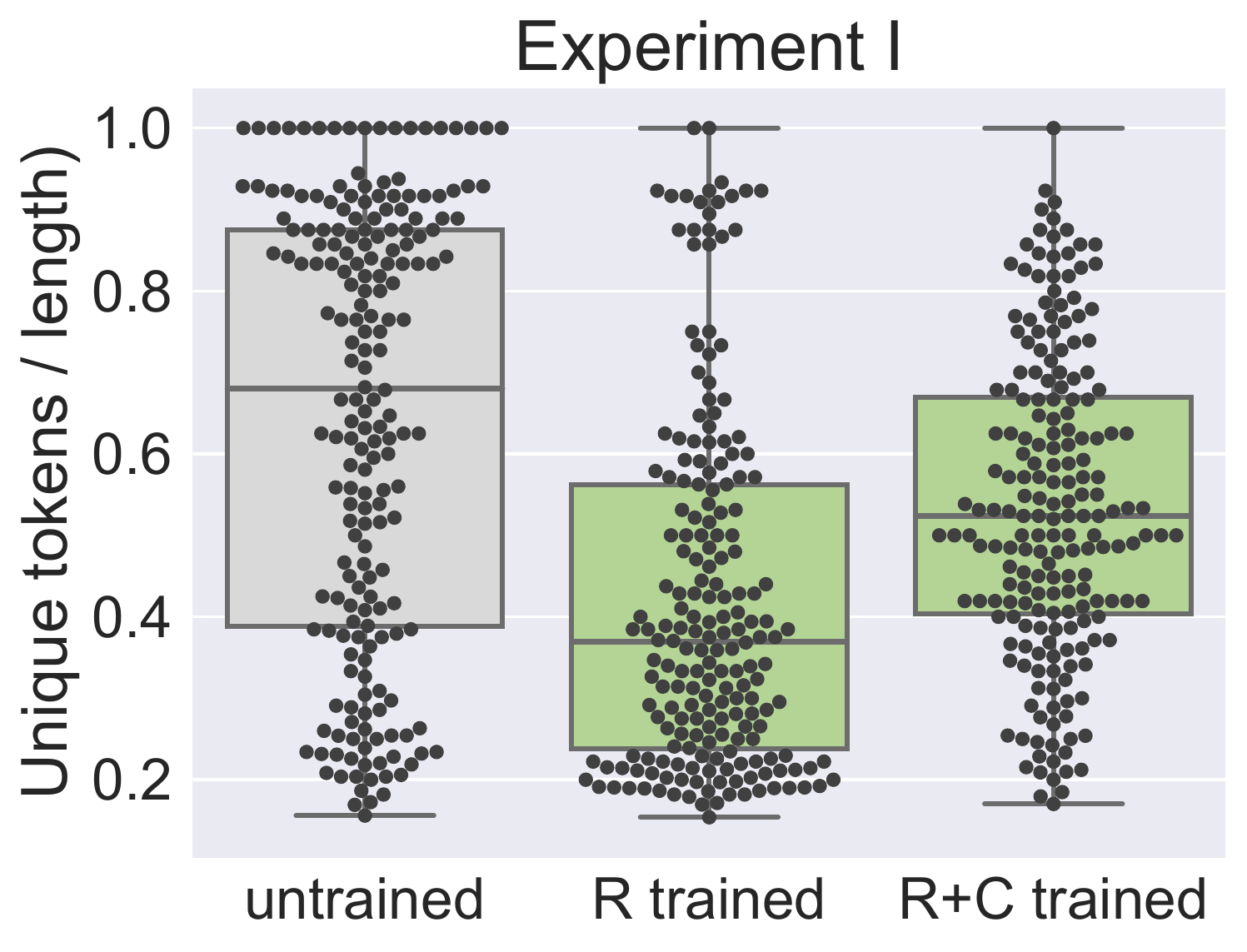}
    \label{fig:exp1_ratio}
    \end{subfigure}
    ~
    \begin{subfigure}[t]{0.48\linewidth}
    \includegraphics[width=\linewidth]{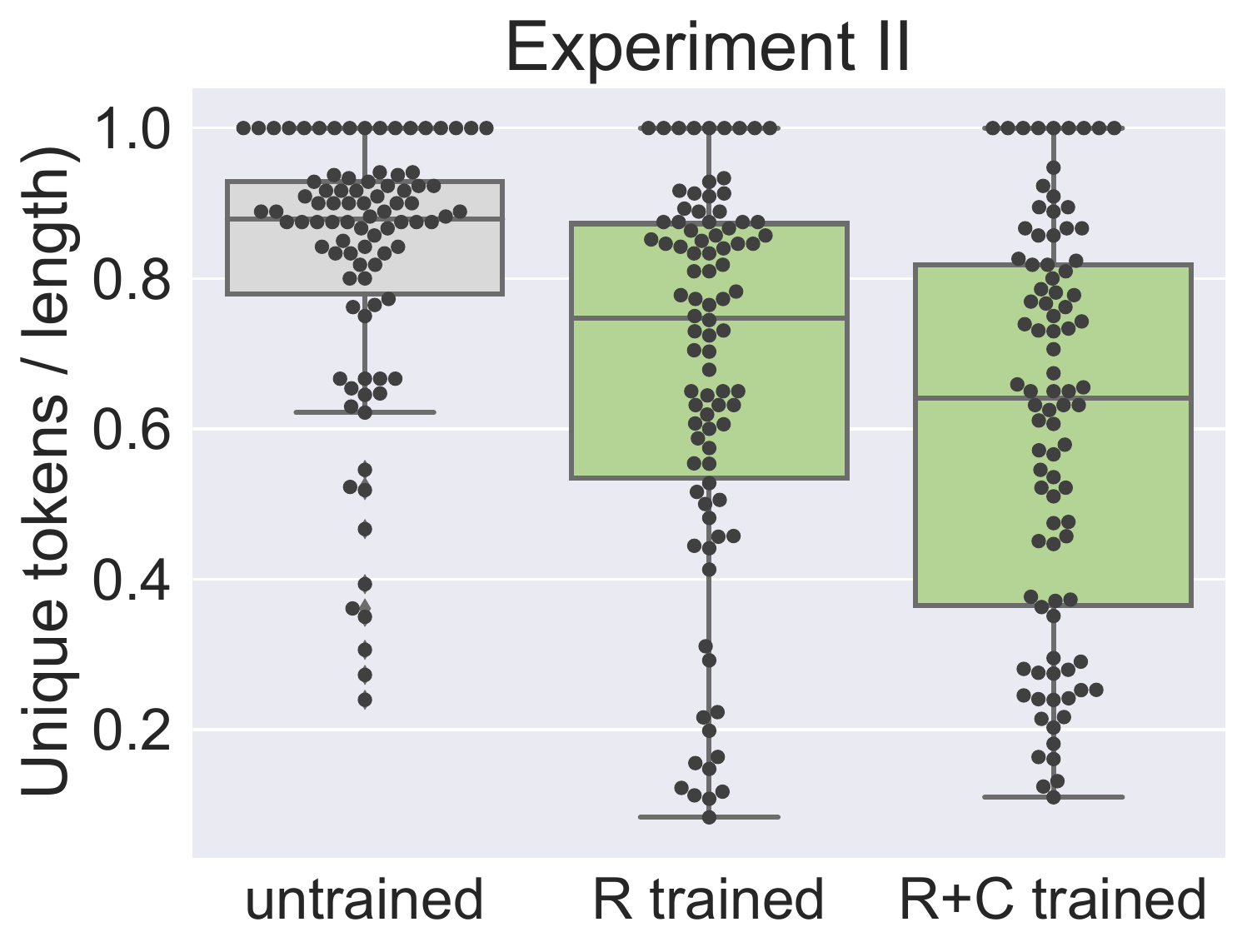}
    \label{fig:exp2_ratio}
    \end{subfigure}

    \caption{Language analysis. \textbf{Top:} After training users in robot capabilities, we observe an increase in absolute ordering language. In Experiment I, we also observe an increase in grouping language after communication training (\S \ref{sec:exp1_design}). \textbf{Bottom:} Word repetition analysis (1.0 = no repetition, 0.0 = complete repetition). We observe that after robot capability training, repetitiveness increases, though this is then offset in Experiment I with communication training. In Experiment II, users overall relied more heavily on algorithmic language due to the complexity of the task.}
    \label{fig:language}
\end{figure}

\subsubsection{Language Analysis}
H1 and H2 supposed that language would change in potentially drastic ways when users are trained on a robot's capabilities.  The left column of Figure \ref{fig:language} shows precisely this effect.  On top, we show how users were split between different linguistic approaches, but once trained they shift almost exclusively to absolute orderings. It is only after we tell them that constraints and partial orderings are acceptable that they begin to change their approach.

In Figure \ref{fig:language} (below), we see a very similar trend in the type token ratios.  Again, untrained users take a diverse set of strategies but once trained they shift towards very repetitive language (low type/token ratios).
Given these jarring changes to the language, we assess their effects on plan efficiencies and correlation with mental demand.

\begin{figure}
    \centering
    \includegraphics[width=0.98\linewidth]{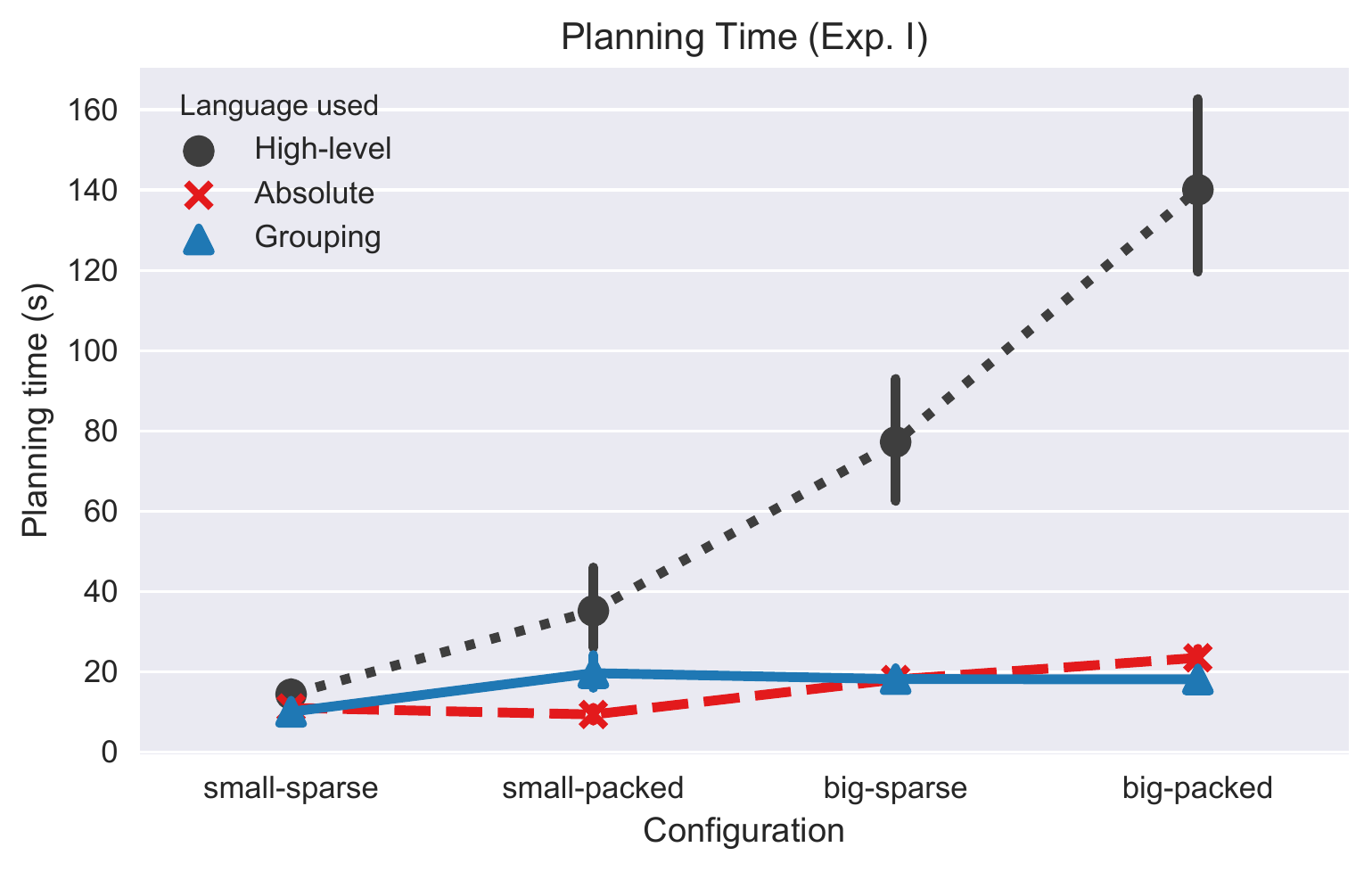}
    \includegraphics[width=0.98\linewidth]{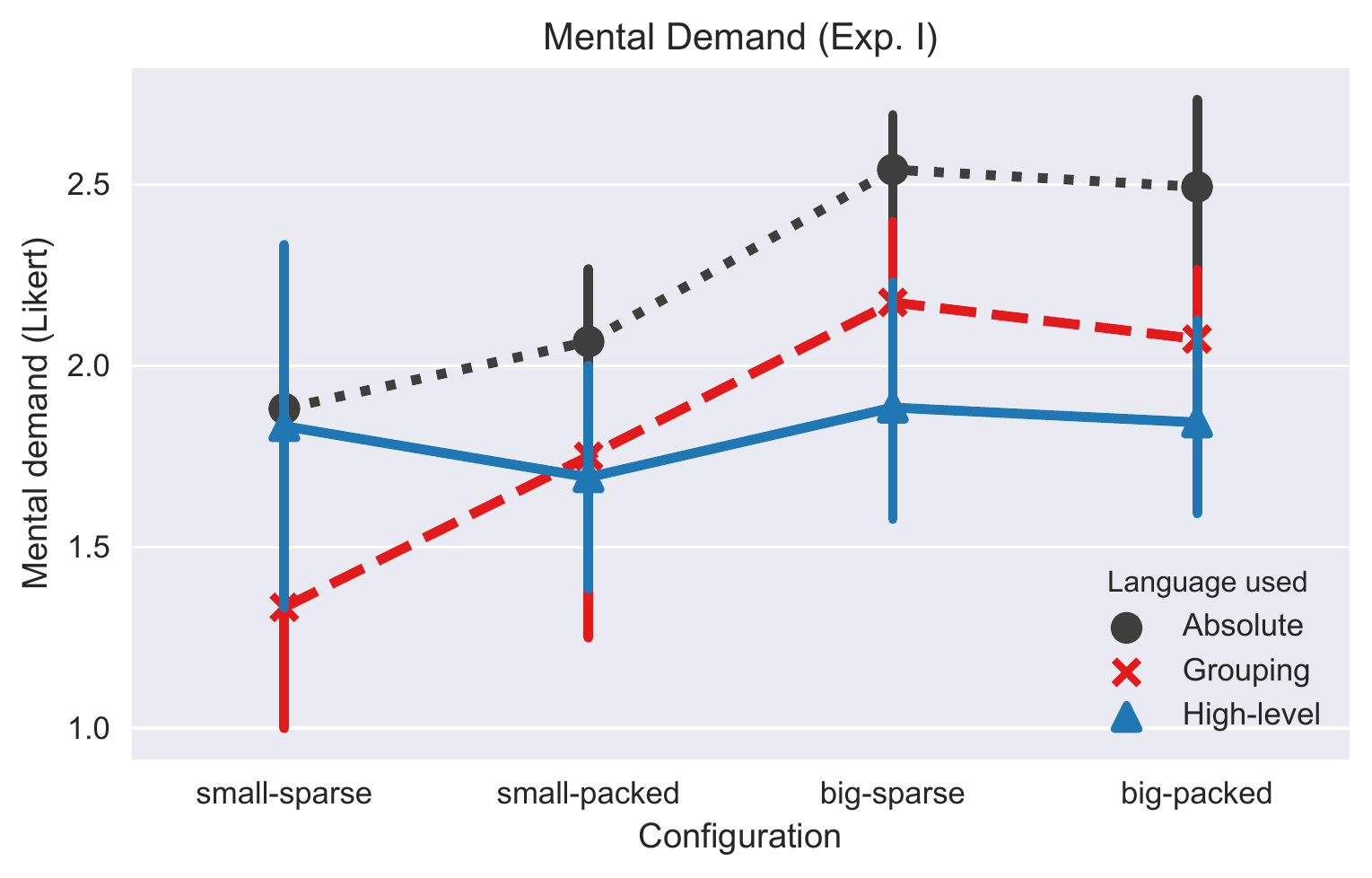}    
    \caption{\textbf{Top:} High-level language does not assist the planner, but both grouping language and absolute orderings provide considerable gains. \textbf{Bottom:} Grouping language strikes a balance in mental demand. Overall, grouping language limits mental demand while benefiting planner performance.}
    \label{fig:mst_exp1}
\end{figure}

\subsubsection{Plan Efficiency vs Mental Demand}
Every type of language (shown in Fig. \ref{fig:configs}) provides an incrementally larger amount of information to the planner.  For every configuration, we ran our planner 10 times on the most common constraint/ordering provided (error bars shown in figure).  Figure \ref{fig:mst_exp1} shows the result of these instructions in planning time.  For analysis, we aggregate configurations into four categories: Packed vs Sparse, and Small ($\leq 5$) vs Large ($\geq$ 6).  When only a high-level plan is provided, we default to the planner's de facto performance.  We use exact sequences for absolute orderings.  Finally, if only a partial ordering is provided, we average over orderings sampled from the set of orderings that lead to geometrically feasible solutions.  Immediately, we see that human insight dramatically speeds up planning.

Next, we plot the mental demand users indicated was required for each type of language and configuration.  First we note that high-level plans are unaffected by task complexity while providing absolute orderings becomes tiring.  Importantly though, our results for grouping language which only partially constrain the task validate H2, as they balance plan efficiency (providing large gains to the planner) while incurring a lower mental demand than fully explicit orderings. 

%
% m/s/t figures
%
\begin{figure*}
    \centering
    \includegraphics[width=0.48\linewidth]{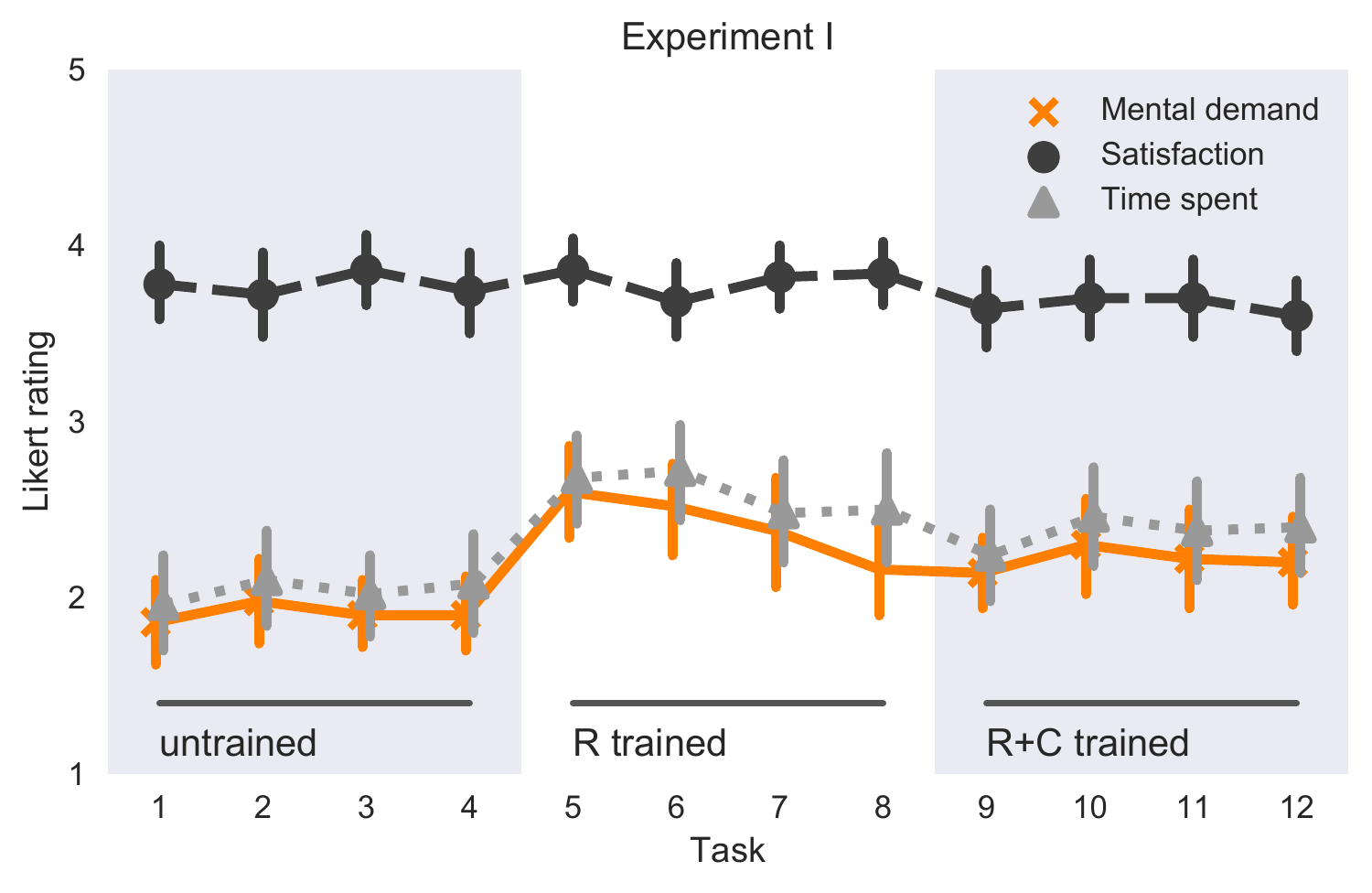}
    \hspace{10pt}
    \includegraphics[width=0.48\linewidth]{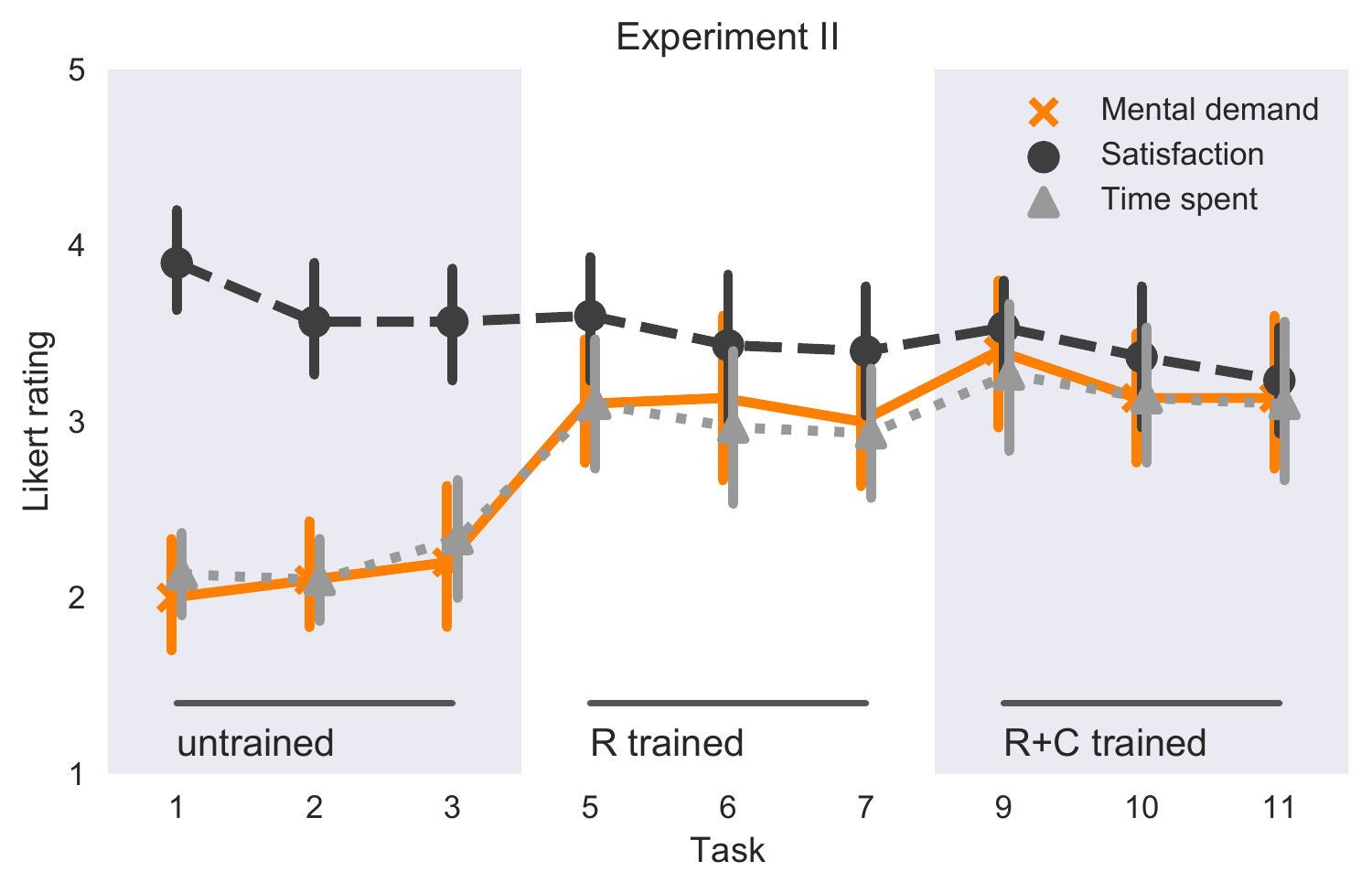}
    \caption{Mental demand, satisfaction, and time ratings for both experiments. Difficulty increases after robot training and language training, but appears to level off as users acclimate to the task.}
    \label{fig:mst}
\end{figure*}

\subsubsection{Mental Exhaustion}
Finally, we investigate how the phases of our task are wearing on the participants.  The top of Figure \ref{fig:mst_exp1} shows how, regardless of the scenario, users are very satisfied with their performance but the time and mental energy they are spending on the task increases and decreases as they become aware of the robot's limitations and linguistic abilities, respectively.  
Their desire to communicate effectively but easily also shows up as a strong preference for scenes in which all cups have a unique color (Figure \ref{fig:exp_preferences}).  This makes sense since naming a color is easier than using spatial language or multiple attributes to disambiguate an object (``red cup'' vs ``red cup on the left'').

\section{Experiment II}
\label{sec:exp2}
Experiment I limited scenarios to at most seven objects to be consistent with most robots+language literature which stay under 15 objects.  In contrast, most natural scenes are more ambiguous with a large number of possible referents.  While complex environments are exponentially more difficult for motion planners, in our simulated environment, we can scale our experiment to see the effects on language.  This leads us to a new hypothesis untested in the literature:\\
\\
\noindent \textbf{H3} \textit{The preference for programmatic language and absolute orderings is an artifact of people's desire to minimize their own cognitive load and therefore an artifact of simple environments.  Specifically, we expect constraint and set based language to be most common in complex environments}.

\subsection{Study Design}
Our design mirrors Experiment I but uses a new set of stimuli which focuses on scaling the number of objects in a scene. These were divided into two scene types with 24 objects in specified locations. In the first, each of the 24 objects was uniquely identifiable by a color with a common name. In the second, each of the 24 objects was randomly assigned a value for each attribute (bottle/cup, blue/red, small/large). Again, we disallow any user overlap from the previous experiment, so all participants are new to the task and untrained.

\subsection{Results}
While the planner will now be too slow to compare as a possible baseline, we still want to investigate how language and mental exhaustion scale to more realistic scenarios.

\subsubsection{Language Analysis}
Where simple environments occasionally elicited explicit ordering language even before training, these are not seen as natural or viable approaches in the richer environment.  Figure \ref{fig:language} (top) shows a dramatic preference for high-level language, and correspondingly, we see very high type/token ratios (bottom).  These are truly natural and diverse instructions.  More importantly, and interestingly, we see a remarkable reluctance to change even after training.  Where knowledge of robot capabilities previously led participants to use the very helpful (though taxing) absolute orderings, now users choose to only provide partial orderings or set-based language. It is only after we explicitly tell them of the types of language we want that absolute orderings become more common. Even then, they are half as common as in the simpler setting.  Again, the lowest type-token ratios in this experiment are higher on average than Experiment I.

\subsubsection{Mental Exhaustion}
Another important differentiation of this result from Experiment I is seen in the changing mental demands.  Figure \ref{fig:mst_exp1} shows how, after training, mental demand increases and never drops.  Where before, we gave participants a ``way out" by suggesting they try and use constraints or orderings to simplify the communication, now they are reluctant to change their language and forcing them to consider plans in detail is highly demanding.  Particularly poignant is that while demand stays constant, satisfaction decreases.  Users are increasingly uneasy about their own instructions and how they will be interpreted.  This leaves open some important questions about how to decide when it is worth pressing the human for their powerful insights if they find the process frustrating.

\begin{figure}
    \centering
    \begin{subfigure}[t]{0.48\linewidth}
    \includegraphics[width=\linewidth]{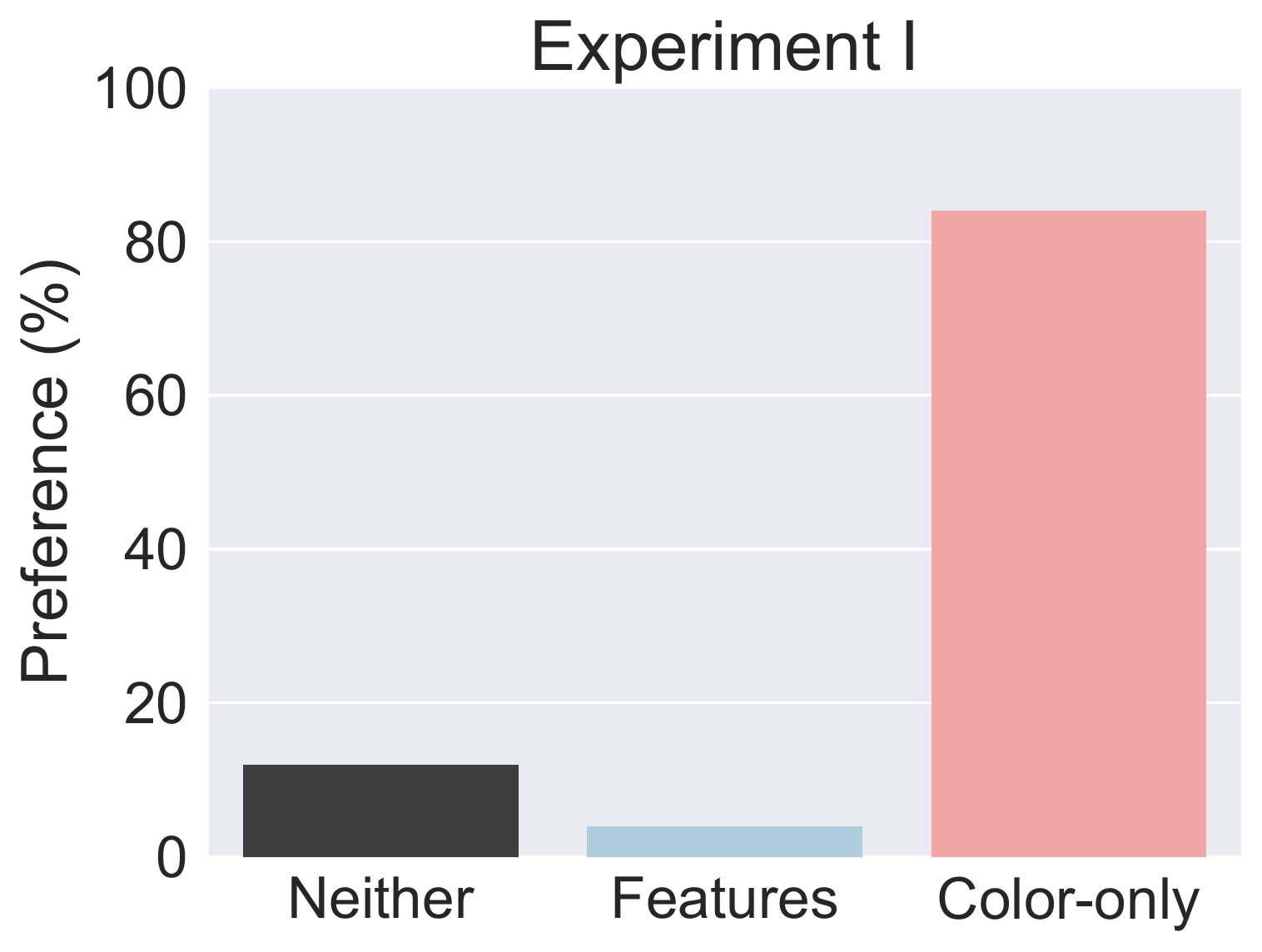}
    %\label{fig:exp1_preferences}
    \end{subfigure}
    ~
    \begin{subfigure}[t]{0.48\linewidth}
    \includegraphics[width=\linewidth]{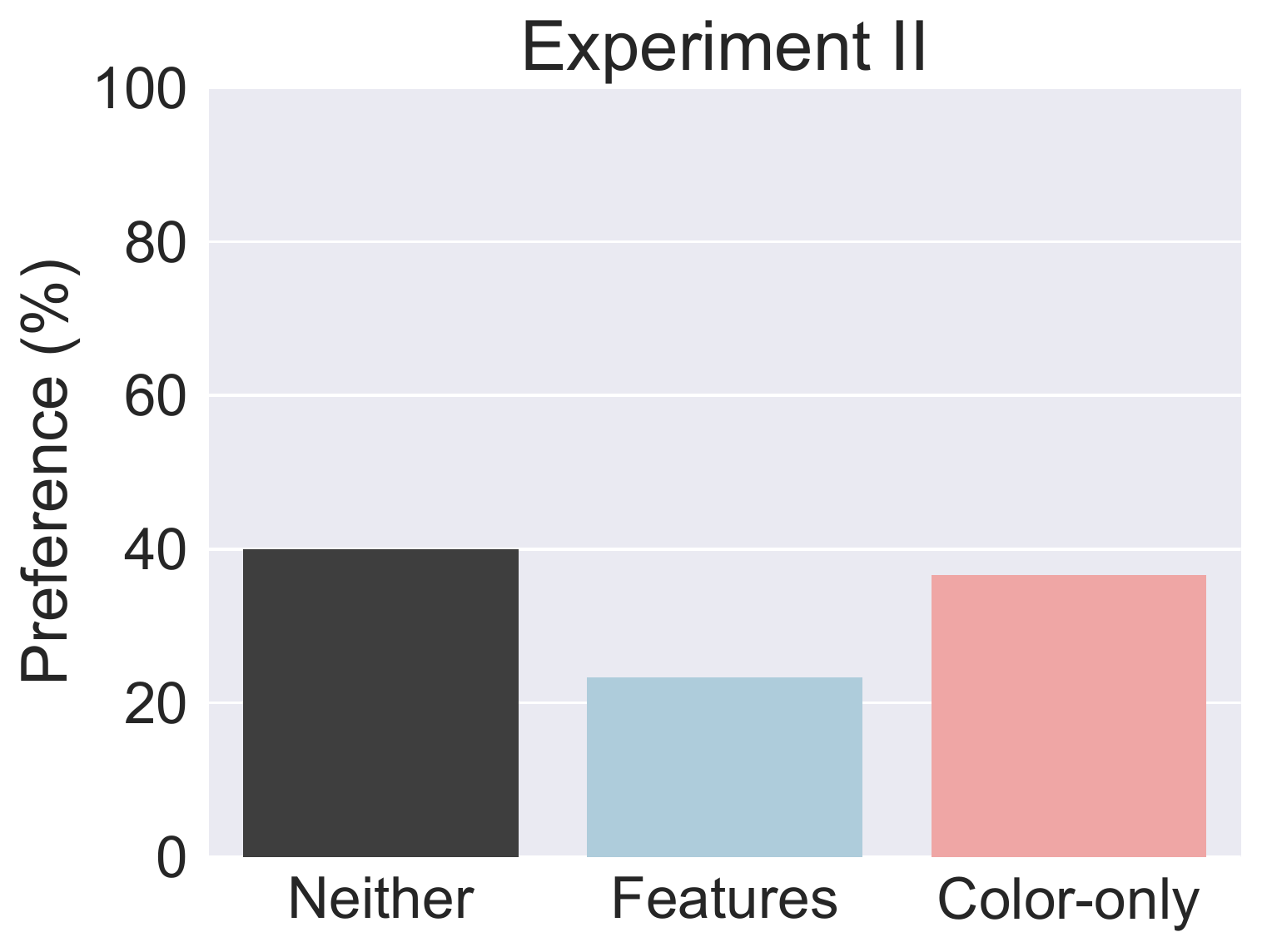}
    %\label{fig:exp2_preferences}
    \end{subfigure}
    \caption{Preference ratings across task configurations for Experiments I and II. In Experiment I, users preferred using simple referring expressions to locate unique objects. Users did not demonstrate any preference in Experiment II.}
    \label{fig:exp_preferences}
\end{figure}

Correspondingly, now that the scenes are complicated, preferences for easily referenced colors versus sets even out.  Figure \ref{fig:exp_preferences} shows that unique colors are still preferred, but there is little agreement as to which setting is actually easiest.

%
% Preferences for both study1 and study2
%

\section{Language Strategies}
\label{sec:lang_strategy}
Upon publication we will make available all of our configurations and the corresponding language for the community to scrutinize, but we include a few examples here in Figure \ref{fig:configs}.  We want to draw particular attention to the different types of high-level language.  Recent results in robotics lead us to believe that many labs can handle absolute orderings and recent work on understanding groups/rows/sets should cover partial orderings, but high-level language appears to be much less homogeneous in nature.  Specifically, in Experiment II the majority of users tried to be helpful by providing us with techniques, strategies, and intuition for the problem.

These example differ from goal-language as they are closer to pseudocode for the correct search procedure.  In our experiments, HERB's abilities nicely parallel a human's arm and so the user might be describing how they would reason about the task.   Equally important for future work is to include details subtle ways the robot differs from their expectations (e.g. HERB's long arms might make close grasps difficult), and then compare the algorithms/heuristics generated by the user.  More technically, we are unaware of any literature that works to interpret and convert these types of hints into planner actions or constraints.

Finally, our participants all tried to be helpful to HERB.  When we compared the most common plan (the ones used for computing motion plans in our plots) to all others produced within the pilot, we saw on very small differences in plan time since nobody strategically instructed HERB to perform infeasible actions.  This may not be true in general for deployed robotics, hinting at a new research question:  How do we detect when a user is being malicious?

\section{Conclusion}
\label{sec:conclusion}

This work discusses the interaction between humans and robots from a language communication perspective.
It investigates the importance of language in shifting autonomy between the human and robot, and when or why a human might choose to be helpful or abdicate responsibility.

We only discussed the language from human to robot in this work, which allows the human to decide how much autonomy they want the robot to exhibit. 
Understanding the variables we introduced helps us calibrate the shared load a human teammate expects.
Analysis and categorization of language also provides insight into how much of the workload the human teammate is willing to take-on or how hard they are working on a given task.
A natural extension is to inquire how a robot should respond (verbally) if they want to strategically ask for help or increase the user's participation to redistribute or optimize the cognitive load.

Humans work very well with each other in teams by communicating goals, plans, heuristics, and asking for help. 
While semantic parsers and Natural Language Processing (NLP) systems may not yet be equipped to handle all of the parsing necessary for the language we have presented, if robots are to serve as helpful team members, we will need to bridge this gap between human language preferences and robot understanding ability.

This is still a young area of research, and language communication in shared autonomy provides an exciting and effective interface to enhance existing sliding autonomy systems~\cite{heger2006sliding} into complex task coordination, where-in task-planning libraries can be extended to be interactive task-planning libraries that elicit language interaction.
We believe these research results will help us to better design agents with bidirectional communication between humans and robots, especially in manipulation tasks, where a robot needs to: (1) precisely understand what a human wants, (2) dynamically monitor the workload distribution, and (3) model the human's characteristic behaviors for optimizing reactions.

\addtolength{\textheight}{-12cm}   % This command serves to balance the column lengths
                                  % on the last page of the document manually. It shortens
                                  % the textheight of the last page by a suitable amount.
                                  % This command does not take effect until the next page
                                  % so it should come on the page before the last. Make
                                  % sure that you do not shorten the textheight too much.

%%%%%%%%%%%%%%%%%%%%%%%%%%%%%%%%%%%%%%%%%%%%%%%%%%%%%%%%%%%%%%%%%%%%%%%%%%%%%%%%

\input{reference.bbl}

\end{document}

%% file: reference.bbl
% Generated by IEEEtran.bst, version: 1.12 (2007/01/11)